\pdfoutput=1

\documentclass[11pt]{article}

\usepackage[preprint]{acl}

\usepackage{times}
\usepackage{latexsym}
\usepackage{enumitem}
\usepackage[T1]{fontenc}

\usepackage[utf8]{inputenc}

\usepackage{microtype}

\usepackage{inconsolata}
\usepackage{adjustbox}
\usepackage{amsmath}
\usepackage{bm}
\usepackage{xcolor}
\usepackage{booktabs}

\usepackage{enumitem}
\usepackage{amssymb}
\usepackage{amsthm}
\usepackage{cleveref}
\usepackage{wrapfig}
\usepackage{stackengine}
\usepackage{comment}
\usepackage{mathtools}

\usepackage{subcaption}
\usepackage{relsize}

\usepackage{float}
\usepackage{rotating}
\usepackage{multirow}
\usepackage{pgfplots}
\usepackage{tikz}
\usepackage{xspace}

\usepackage{mathabx}
\definecolor{cmzhao}{rgb}{0.1, 0.8, 0.1}

\theoremstyle{plain}

\theoremstyle{remark}

\definecolor{LightSteelBlue1}{RGB}{202,225,255}
\definecolor{LightPink}{RGB}{245,191,210}
\definecolor{Moccasin}{RGB}{255, 228, 181}

\definecolor{LightSteelBlue1}{RGB}{202,225,255}
\definecolor{LightPink}{RGB}{245,191,210}
\definecolor{Moccasin}{RGB}{255, 228, 181}
\setenumerate[1]{itemsep=0pt,partopsep=0pt,parsep=\parskip,topsep=0pt}
\setitemize[1]{itemsep=0pt,partopsep=0pt,parsep=\parskip,topsep=0pt}
\setdescription{itemsep=0pt,partopsep=0pt,parsep=\parskip,topsep=0pt}
\title{High-Rank Structured Modulation for Parameter-Efficient Fine-Tuning}

\author{Yongkang Liu$^{1}$ Xing Li$^{1}$ Mengjie Zhao$^{1}$ Shanru Zhang$^{1}$ Zijing Wang$^{1}$ \\
  \textbf{Qian Li$^{3}$ Shi Feng$^{1}$ Feiliang Ren$^{1}$ Daling Wang$^{1}$ and Hinrich Schütze$^{2,4}$} \\
        $^1$Northeastern University, China;
        $^2$CIS, LMU Munich, Germany \\
        $^3$Shandong University, China\\
        $^4$Munich Center for Machine Learning (MCML), Germany \\
        \texttt{misonsky@163.com}
}

\begin{document}
\maketitle
\begin{abstract}
As the number of model parameters increases, parameter-efficient fine-tuning (PEFT)
has become the go-to choice for tailoring pre-trained large language models. 
Low-rank Adaptation 
(LoRA) uses a low-rank update method to simulate full parameter fine-tuning, which is widely 
used to reduce resource requirements. However, decreasing the rank encounters challenges with 
limited representational capacity when compared to full parameter fine-tuning. 
We present \textbf{SMoA},
a high-rank \textbf{S}tructured \textbf{MO}dulation \textbf{A}dapter that 
uses fewer trainable parameters while maintaining
a higher rank, thereby improving the model's representational capacity and offering improved 
performance potential. 
The core idea is to freeze the original pretrained weights and selectively amplify or suppress important features 
of the original weights across multiple subspaces. The subspace 
mechanism provides an efficient way to increase the capacity and complexity of a model. We conduct
both theoretical analyses and empirical studies on various tasks. 
Experiment results show that SMoA outperforms LoRA and its variants on 10 tasks, with extensive ablation 
studies validating its effectiveness.
\end{abstract}

\section{Introduction}

Large language models (LLMs)~\cite{touvron2023llama,zhang2022opt,achiam2023gpt} have 
demonstrated remarkable performance improvements across a wide range of natural language 
processing tasks. Fine-tuning (FT) is a prevailing way for tailoring LLMs for specific
downstream tasks~\cite{liu2025look}. However, the vast scale of LLMs makes full-parameter FT 
prohibitively expensive, especially in resource constrained environments. Parameter-Efficient 
Fine-Tuning (PEFT) is proposed to alleviate the high memory requirements~\cite{lester2021power,liu2024gpt,hu2022lora,hayou2024lora+}.
PEFT achieves low-memory FT by reducing the number of trainable parameters, such as prompt tuning~\cite{lester2021power},
adapter tuning~\cite{rebuffi2017learning}.

\begin{figure}[t]
    \centering
    \includegraphics[width=0.90\columnwidth]{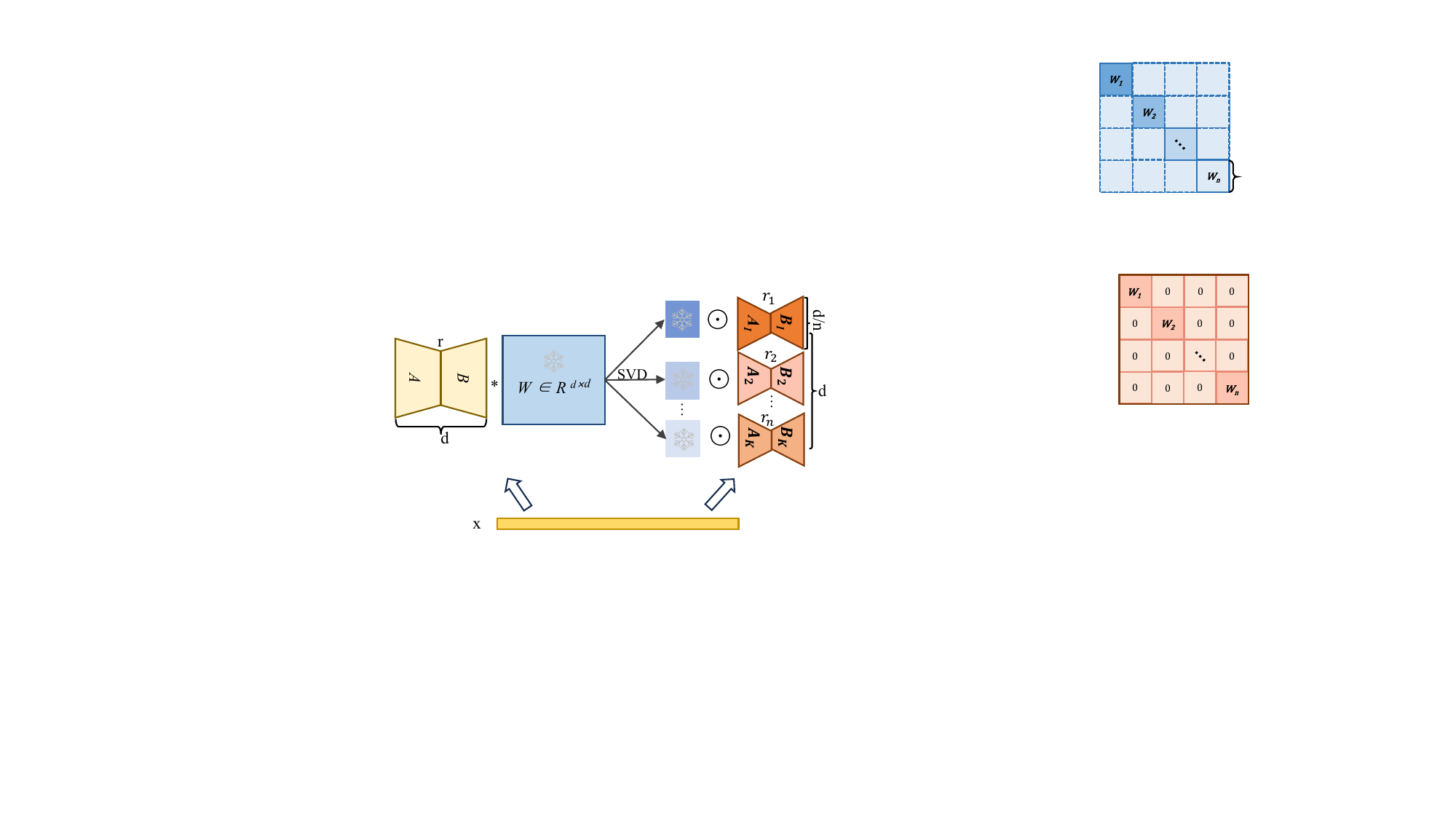}
  \caption{Comparison between LoRA (left) and and the proposed SMoA (right). The core idea of SMoA lies in diversifying the modulation of the original weights across multiple subspaces.}
  \label{fig:hime_architecture}
\end{figure}
Low-rank adaptation~\cite{hu2022lora}, as a classic example of PEFT, is widely being used due to its extremely 
small additional memory overhead and because it comes without additional inference latency. However,
previous studies~\cite{jiang2024mora, liu2024dora, zhuang2024time} have shown that LoRA and most 
of its variants~\cite{lialin2023relora, hayou2024lora+} struggle with complex tasks, such as 
intricate mathematical reasoning and learning new knowledge~\cite{liu2025look}. As shown in 
Figure~\ref{fig:hime_architecture}, the update matrix $\Delta W$ of LoRA is constructed as the product of two 
low-rank matrices, $A \in \mathbb{R}^{d \times r}$ and $B \in \mathbb{R}^{r \times d}$, resulting in an update with 
rank at most $r$. The low-rank property limits the model's representational capacity. 

A natural solution to this issue is to increase the rank of the updated parameter matrix to 
increase
its capacity. 
Most existing methods mitigate this problem by employing multiple parallel 
LoRAs. ReLoRA~\cite{lialin2023stack} and COLA~\cite{xia2024chain} progressively merge old LoRA to pre-train weights and stack new LoRAs 
during training. MELoRA~\cite{ren2024melora} increases the effective rank of the model by stacking multiple 
mini LoRA modules. However, there may be overlap in parameters between the LoRA module series.
HiRA~\cite{huang2025hira} improves the effective rank of the model by leveraging the original 
weights through a Hadamard product with the LoRA module. Its effective rank remains 
constrained by the rank of the LoRA module.

In this paper, we propose a simple yet effective method, called structured modulation 
adapters (SMoA), that performs structured modulation of the main features of the original weights
in multiple subspaces, as shown in Figure 1 (right). We demonstrate theoretically that
SMoA ensures a higher rank \emph{without imposing an additional parameter overhead}, which allows the 
use of LoRA modules with different ranks in each subspace. Specifically, SMoA achieves 
high-rank, structured, and stable task adaptation of model weights with 
controllable parameter overhead by introducing multiple low-rank, decoupled Hadamard multiplicative 
LoRA modules within the principal singular subspace of the pre-trained model.
Each LoRA module is constrained to a distinct singular subspace, thereby preventing overlap and reducing homogenization across modules.
We conduct extensive experiments across diverse tasks and models to demonstrate the efficacy
of SMoA. Evaluations are performed using Llama-2-7B~\cite{Touvron2023Llama2O} and Llama-3-8B~\cite{Dubey2024TheL3}.
on commonsense reasoning, 
dialogue generation and 
mathematical reasoning tasks.
Results indicate that SMoA achieves state-of-the-art performance compared to most baselines.

We summarize our contributions as follows:

\begin{itemize}[leftmargin=*]
    \item We propose a new method (SMoA) on top of LoRA that makes it achieve a higher rank and better performance without imposing an additional parameter overhead.
    \item We theoretically demonstrate that SMoA maintains a higher and more flexible rank than LoRA and its variants.
    \item Experiment results show that SMoA achieves state-of-the-art results compared to LoRA and its variant on multiple different benchmarks.
\end{itemize}

\section{Related work}
The extraordinary scale and computational cost of fine-tuning modern LLMs have made PEFT 
a critical research area. The core objective of PEFT is to achieve performance comparable 
to full fine-tuning by updating only a minimal subset of model parameters~\cite{liu2025look}.
Existing PEFT methods are generally divided into four categories: 
prompt-based methods, 
adapter methods, 
selection-based methods, 
and low-rank adaptation, 
among which low-rank adaptation constitutes a major category~\cite{han2024parameter}. 
LoRA~\cite{hu2022lora} 
factorizes the weight updates into two low-rank matrices, achieving parameter-efficient 
adaptation without incurring additional inference cost. 
VeRA~\cite{kopiczko2023vera} extends LoRA by freezing random matrices and learns only a small number of scaling 
vectors. DoRA~\cite{liu2024dora} decomposes weight updates into direction and magnitude components, explicitly modeling 
scale changes to improve adaptation performance. FLoRA~\cite{wenbatched} proposes an example-specific 
adapter based on the Hadamard product. 
SSMLoRA~\cite{yu2025ssmlora} extends LoRA by incorporating a state-space model 
to connect low-rank adapters across layers, maintaining performance with sparser insertions.

Different from low-rank updates, MoRA~\cite{jiang2024mora} replaces standard low-rank updates 
with a square “higher-rank” matrix, reducing parameters and computation while maintaining 
expressive capacity. MELoRA~\cite{ren2024melora} ensembles multiple small low-rank adapters to improve model 
performance and robustness without imposing extra parameters. 
HiRA~\cite{jiang2024mora} uses 
element-wise products to enable “high-rank” task-specific weight updates to improve expressive capacity. 
CoTo~\cite{zhuang2025come} gradually increases adapters’ activation probability during fine-tuning, encouraging broader 
loss-landscape exploration and balanced optimization to improve generalization and training efficiency compared with standard LoRA.
Different from existing methods that rely on complex static compression and merging multiple LoRA modules, 
SMoA improves the rank of the model by adapting the principal singular components of the original model 
to multiple subspaces in a diverse manner.

\section{Methodology}
\label{sec:method}
~\citet{liu2025look} demonstrates that PEFT fine-tuning primarily activates existing 
capabilities of the original LLM rather than introducing new knowledge. As a result, performance 
gains from PEFT are largely attributable to better utilization and steering of pre-existing knowledge. 
This finding suggests that PEFT is particularly effective for adaptation scenarios where 
relevant knowledge is already present in the backbone model. We provide a formal description of this phenomenon.
Let a pretrained language model be defined as: $f_{\theta_0}: \mathcal{X} \rightarrow \mathcal{Y}
$, where $\theta_0$ denotes the parameters learned during large-scale pretraining, implicitly 
encoding 
rich 
linguistic knowledge and reasoning capabilities.

From a representation learning perspective, PEFT can be formalized as learning a constrained 
parameter perturbation:
\begin{equation}
    \theta = \theta_0 + \Delta\theta, \quad \Delta\theta \in \mathcal{S}
\end{equation}
where $\mathcal{S}$ is a low-dimensional, structured subspace determined by the specific 
PEFT method (e.g., LoRA), and the pretrained parameters $\theta_0$ remain frozen. In functional terms, this can be expressed as:
\begin{equation}
\label{eq:2}
    f_{\theta_0 + \Delta\theta} \approx g_{\Delta\theta} \circ f_{\theta_0}
\end{equation}
where $g_{\Delta\theta}$ denotes a lightweight transformation that modulates intermediate 
representations produced by the backbone model, $\circ$ denotes function composition.
Under this formulation, PEFT does not explicitly alter the underlying knowledge representations 
encoded in $f_{\theta_0+ \Delta\theta}$. Instead, it primarily re-weights, activates, or 
suppresses existing representational dimensions, thereby improving the accessibility and 
utilization of pretrained knowledge for some specific downstream tasks. 

Under this formulation, the lightweight transformation $g_{\Delta\theta}$ can be instantiated 
as a linear residual modulation applied to intermediate representations, which can be formalized as:
\begin{equation}
\label{eq:3}
    g_{\Delta \theta}(h) = h + \Delta W x
\end{equation}
$f_{\theta_0}(\cdot)$ can be represented as:$f_{\theta_0}=W_0x$. Equation~\ref{eq:2} can be 
specified as follows:
\begin{equation}
\label{eq:4}
    f_{\theta_0 + \Delta\theta} \approx g_{\Delta\theta} \circ f_{\theta_0}=g_{\Delta\theta}(f_{\theta_0}(x))
\end{equation}
For a pre-trained weight $W_0$, combining equations~\ref{eq:3} and ~\ref{eq:4} composite function can be formalized as:
\begin{equation}
    f_{\theta_0 + \Delta\theta}(x) = W_0 x + \Delta W x
\end{equation}

LoRA, where $\Delta W=AB$, can be viewed as performing parameter updates within a constrained 
adaptation space, where the weight updates are restricted to lie on a low-rank manifold, 
thereby reducing the intrinsic rank of the learned updates. A known limitation of LoRA is 
that the learned weight updates often concentrate on a small number of dominant singular 
directions, potentially limiting the rank of $\Delta W$ and expressiveness.

\subsection{Structured Modulation Adapter}
Different from LoRA, our proposed
SMoA ensures that every adaptation subspace corresponds to a comparable 
amount of pretrained representational energy. This enables different parameters to focus on 
distinct singular directions, thereby enhancing the representational capacity and the rank of 
model. First, SMoA, applies singular value decomposition (SVD):
\begin{equation}
   W_0 = U \Sigma V^\top, \quad
    \Sigma = \mathrm{diag}(\sigma_1, \dots, \sigma_{d})
\end{equation}
where $\sigma_1 \ge \cdots \ge \sigma_{d}$ and $d$ is the number of singular values. 
The singular values $\{\sigma_i\}$ encode the 
relative importance of the various pretrained representational directions.
We define the cumulative spectral energy function as
\begin{equation}
    E(i) = \frac{\sum_{j=1}^{i} \sigma_j}{\sum_{j=1}^{d} \sigma_j}
\end{equation}

Given $K$ adaptation subspaces, we deterministically partition the singular directions into $K$ 
disjoint subsets by evenly dividing the cumulative spectral energy. Specifically, the index set assigned 
to the $k$-th LoRA subspace is defined as:
\begin{equation}
\label{eq:8}
    \mathcal{I}_k =\left\{i \;\middle|\;\frac{k-1}{K} < E(i) \le \frac{k}{K}\right\}
\end{equation}
This construction ensures that each subspace is responsible for approximately the same 
amount of pretrained spectral energy, without introducing any additional hyperparameters.
For the $k$-th adaptation subspace, we introduce a LoRA module parameterized by $A_k$ and $B_k$. 
We construct a fixed, non-learnable spectral modulation tensor:
\begin{equation}
\tilde{\Sigma}_k = U \mathrm{diag}\left(I_{\mathcal{I}_k} \odot \sigma \right) V^\top
\end{equation}
where $I_{\mathcal{I}_k}$ denotes the indicator vector of $\mathcal{I}_k$. The effective update of the
$k$-th subspace is then given by:
\begin{equation}
    \widehat{\Delta W}_k= (B_k A_k) \odot \tilde{\Sigma}_k
\end{equation}
The adapted weight matrix is obtained as:
\begin{equation}
    W = W_0 + \mathrm{Concat} \big(\widehat{\Delta W}_1,\,\dots,\,\widehat{\Delta W}_K\big)
\end{equation}

\subsection{Matrix Rank Theory}
\label{sec:32}
Given the same parameter budget, we theoretically demonstrate that SMoA attains a superior 
rank ratio compared to LoRA. Suppose the parameters of the LoRA module are $A \in \mathbb{R}^{d \times r}$ and $B \in \mathbb{R}^{r \times d}$, 
with rank bounded by $r$. In SMoA, for the $k$-th adaptation subspace, the rank of $A_kB_k$ is as follows:
\begin{equation}
\text{rank}(A_kB_k) = \text{min}(\text{rank}(A_k),\text{rank}(B_k)) \le r/K
\end{equation}   
where $A_k\in \mathbb{R}^{(\frac{r}{K}\times\frac{d}{K})}$,
$B_k\in \mathbb{R}^{(\frac{r}{K}\times\frac{d}{K})}$, $r \ll d$.
For $\tilde{\Sigma}_k$, its rank can be expressed as:
\begin{equation}
    \text{rank}(\tilde{\Sigma}_k) = |\mathcal{I}_k| = I_k
\end{equation}
A property of the Hadamard product is that:
\begin{equation}
   \text{rank} (P \odot Q) \le \text{rank} (P) \times \text{rank} (Q)
\end{equation}
According to the above theorem, the upper bound of the rank of $\widehat{\Delta W}_k$ is:
\begin{equation}
   \text{rank} (\widehat{\Delta W}_k) \le I_k \times \frac{r}{K}
\end{equation}
The upper bound of the rank of $\Delta W$is:
\begin{equation}
\begin{split}
    \text{rank}(\Delta W) &= \sum_i^n\text{rank}(\widehat{\Delta W}_i) \\
    &\le \sum_i^K (I_i \times \frac{r}{K}) \\
    &= \frac{r}{K} \times \sum_i^K I_i
\end{split}
\end{equation}
where $\Delta W = \mathrm{Concat} \big(\widehat{\Delta W}_1,\,\dots,\,\widehat{\Delta W}_K\big)$.
$K$ represents the number of subspaces, under the same parameter conditions, we have $K\le r$.
$\tilde{\Sigma}_k$ is a singular space partition of the same weight matrix, therefore:
\begin{equation}
\label{eq:17}
    \cup_{k=1}^K I_k\subseteq \{1,...,d\}
\end{equation}
According to Equation~\ref{eq:17}, we have $\sum_i^K I_i \le Kd$. The rank of $\Delta W$ is:
\begin{equation}
\begin{split}
    \text{rank}(\Delta W) &\le \frac{r}{K} \times \sum_i^K I_i \\
    &\le \frac{r}{K} \times Kd \\
    &\le r \times d
\end{split}
\end{equation}
When $\sum_i^K I_i = K$, the upper bound of the rank of $\Delta W$ degenerates to the standard LoRA.
Compared to LoRA, SMoA has the following advantages:
\paragraph{(1) SMoA maintains a higher rank with fewer parameters.}
According to Section~\ref{sec:32}, SMoA significantly raises the theoretical upper bound on the model rank.
This demonstrates the capability of SMoA to achieve a higher rank with fewer parameters.
In SMoA, each subspace-specific 
LoRA module is responsible for a distinct singular subspace, thereby avoiding overlap across subspaces. Simple concatenation is equivalent to block-diagonal concatenation,
and in fact, we also reorganize the parameters in a block-diagonal manner, which guarantees that the rank of $\Delta W$
equals the sum of the ranks of the individual subspaces. In SMoA, the trainable parameters for the $K$ subspaces are:
$K \times (\frac{d}{K} \times \frac{r}{K} + \frac{r}{K} \times \frac{d}{K})=\frac{2dr}{K}$, and the corresponding rank upper bound is $r \times d$.
Under the same parameter budget, standard LoRA admits an upper bound of $\frac{r}{K}$ on the achievable rank,
corresponding to LoRA parameters $A$ and $B$ having shapes of $d\times(\frac{r}{K})$ and $(\frac{r}{K}) \times d$, respectively.
Clearly, the potential rank of SMoA is $K\times d$ times higher than that of LoRA.

\paragraph{(2) SMoA has a flexible rank strategy.} 
The model's rank can be improved without increasing the number of parameters.
In SMoA, each subspace admits an independent rank, which allows different subspaces to have different rank sizes.
We can set the rank of each subspace LoRA to $r$. So individual LoRA modules denoted as A in 
$\mathbb{R}^{r\times \frac{d}{K}}$ and B in $\mathbb{R}^{\frac{d}{K}\times r}$ are configured,
with a total count of trainable parameters being $2\times r\times d$ and the equivalent rank range is
from $K\times r$ to $K\times r \times d$. Adjusting the hyperparameter $K$ allows for modulation of the
equivalent rank without necessitating an increase in the overall parameter count.
\begin{table*}[ht]
\begin{adjustbox}{max width=0.9\textwidth, center}
\includegraphics[width=\textwidth]{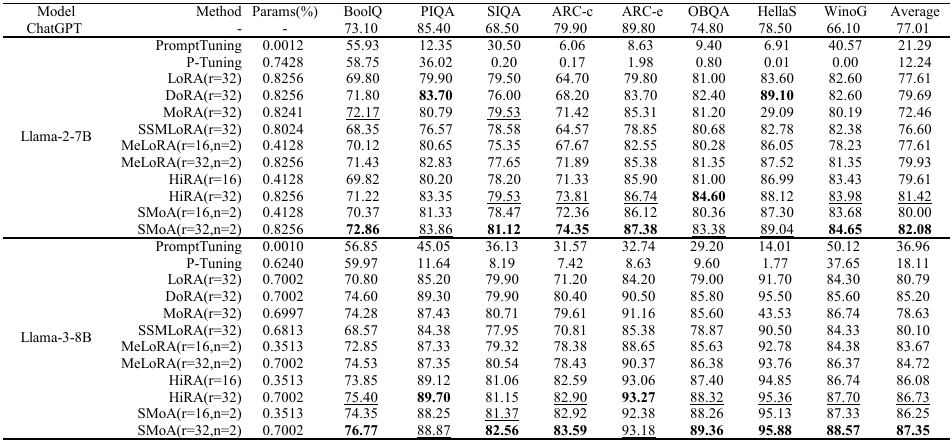}
\end{adjustbox}
\caption{Accuracy comparison among various PEFT methods on commonsense reasoning datasets. Results for 
ChatGPT are sourced from~\citep{liu2024dora}. The best performance within each LLM is indicated in \textbf{bold},
while the second best performance is highlighted in \underline{underline}.}
\label{tab:commonsense}
\end{table*}
\section{Experiments}
We conduct experiments on the commonsense reasoning, dialogue generation, and mathematical reasoning datasets\footnote{Please refer to Appendix~\ref{app:data} for a detailed description}.

\label{sec:experiments}

\begin{table*}[ht]
\begin{adjustbox}{max width=0.9\textwidth, center}
\includegraphics[width=\textwidth]{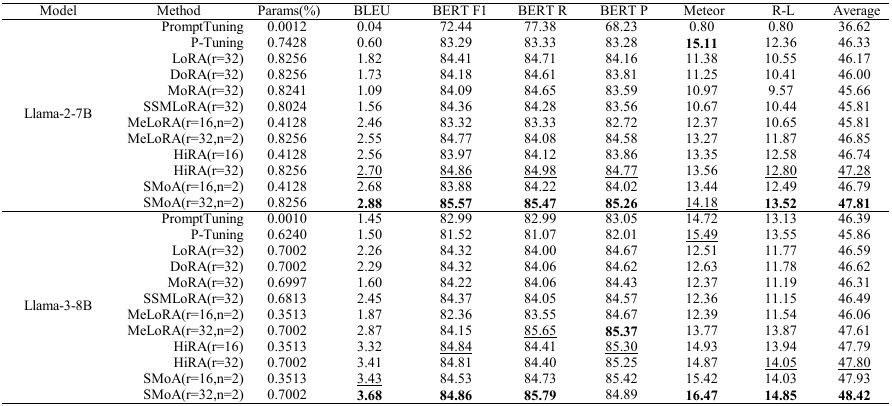}
\end{adjustbox}
\caption{Performance comparison among various PEFT methods on the CONVAI2 dataset, where
BERT F1, BERT-R, and BERT-P denote the F1, Precision, and Recall based on the BERT score, respectively.}
\label{tab:results2}
\end{table*}
\subsection{Baselines}
We compare \textsc{SMoA} with a set of representative PEFT methods.
\begin{itemize}[leftmargin=*]
    \item \textbf{Prompt Tuning}~\cite{lester2021power} adapts pretrained language models by 
    learning a small set of trainable prompt embeddings  while keeping the model parameters 
    frozen, enabling efficient task adaptation with minimal additional parameters.
    \item \textbf{P-Tuning v2}~\cite{liu2021ptuning} is an advanced prompt-based learning method that injects trainable continuous prompts into every layer of a pretrained language model, enabling parameter-efficient tuning that approaches full fine-tuning performance, especially for large models.
    \item \textbf{LoRA}~\cite{hu2022lora} is a parameter-efficient fine-tuning method that adapts large pretrained models by inserting low-rank trainable matrices into existing weight updates, achieving strong performance while updating only a small number of parameters.
    \item \textbf{DoRA}~\cite{liu2024dora} is a parameter-efficient fine-tuning method that improves upon LoRA by decomposing weight updates into magnitude and direction, enabling more expressive adaptations while maintaining low trainable parameter counts.
    \item \textbf{MoRA}~\cite{jiang2024mora} is a parameter-efficient fine-tuning method that extends LoRA by using a mixture of multiple low-rank adaptation modules, allowing the model to capture more diverse and complex task-specific behaviors with minimal additional parameters.
    \item \textbf{SSMLoRA}~\cite{yu2025ssmlora} is a parameter-efficient fine-tuning method that enhances LoRA by incorporating state space model (SSM)–based structured adaptations, improving the modeling of long-range dependencies while keeping the number of trainable parameters low.
    \item \textbf{MeLoRA}~\cite{ren2024melora} is a parameter-efficient fine-tuning method that improves LoRA by reducing activation and optimization memory costs, enabling efficient training and adaptation of large language models under limited hardware resources.
    \item \textbf{HiRA}~\cite{huang2025hira} is a parameter-efficient fine-tuning method that extends LoRA by organizing low-rank adaptations in a hierarchical manner, allowing models to flexibly capture both coarse- and fine-grained task-specific knowledge with minimal additional parameters.
\end{itemize}

\begin{figure}[t]
    \centering
    \includegraphics[width=0.9\columnwidth]{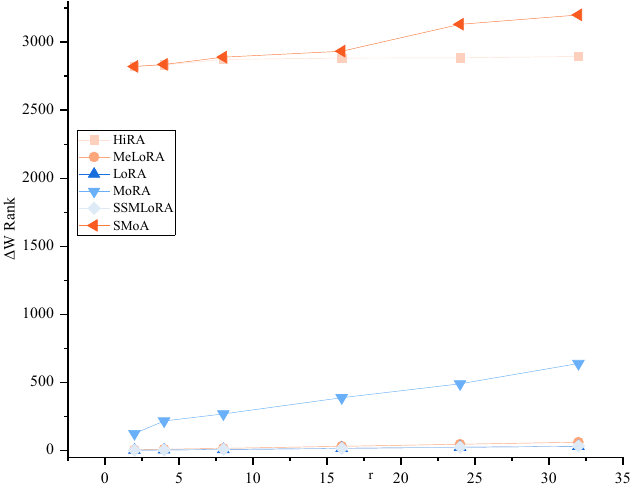}
  \caption{Comparison of the rank of incremental weight $\Delta W$ for different PEFT methods. For MeLoRA and SMoA, the number of LoRA modules is set to 2.}
  \label{fig:rank}
\end{figure}
\subsection{Evaluation Metric}
We employ accuracy as the primary metric for BoolQ, PIQA, SIQA, ARC-c, ARC-e, 
OBQA, HellaS and WinoG. For each test instance, the language models generate answers based on 
the provided queries. We then check for the presence of specific answer 
keywords (e.g., "true" or "false" for BoolQ). The first occurrence of such a keyword is recorded 
as the model’s answer. If no relevant keyword is found, the model is deemed to have failed in 
answering the commonsense reasoning question correctly. This method ensures consistent 
evaluation of the model's performance across all eight tasks~\cite{hu2023llm,liu2024dora}.
For the CONVAI2 dataset, we use the BLEU~\cite{papineni2002bleu} and BERT Score~\cite{zhang2019bertscore} 
as the evaluation metrics.
\subsection{Implementation Details}
Following the identical training setup to~\cite{huang2025hira}, except learning rate adjustments.
We implement SMoA on the Llama-2-7B and Llama-3-8B models with $r=16$ and $r=32$, respectively.
We use the Adamw optimizer~\cite{loshchilov2017decoupled} with a learning rate of 0.001 and 
warms up of 1000 steps.
For BoolQ, PIQA, SIQA, ARC-c, ARC-e, OBQA, HellaS and WinoG datasets, we fine-tune LLMs for 5 
epochs, with evaluations at every 100 step to select the best checkpoint based on validation set.
For LoRA, DoRA, MoRA, HiRA, SSMLoRA, MeLoRA and SMoA methods, we set learnable incremental parameters 
for query, key, value, two linear layers, and down and up projection layers in attention modules.
To ensure fair comparisons, we maintain an equivalent or comparable number of trainable parameters. 
For Prompt Tuning and P-Tuning, which naturally involve fewer trainable parameters due to their use of 
prefix soft prompts, we make necessary adjustments to ensure the number of trainable parameters 
remains comparable. Hyperparameter configuration of SMoA is shown in Table~\ref{tab:smoa_hyperparameter} (Appendix~\ref{app:hyperparameter}).
The general hyperparameter configurations are shown in Table~\ref{tab:gene_hyparameters} (Appendix~\ref{app:hyperparameter}).
\begin{table}[ht]
\begin{adjustbox}{max width=0.85\columnwidth, center}
\includegraphics[width=\columnwidth]{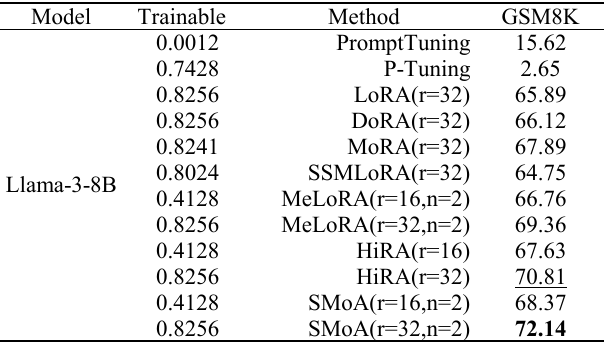}
\end{adjustbox}
\caption{Performance comparison among various PEFT methods on mathematical reasoning task.}
\label{tab:gsm8k}
\end{table}
\section{Results \& Analysis}
\subsection{Commonsense Reasoning Tasks}
Table~\ref{tab:commonsense} reports the accuracy comparison of various PEFT methods on a 
diverse set of commonsense reasoning benchmarks, including BoolQ, PIQA, SIQA, ARC-c, ARC-e, OBQA, 
HellaSwag, and WinoGrande, evaluated on LLaMA-2-7B and LLaMA-3-8B backbones. 
With both backbones, our proposed SMoA (r=32, n=2) achieves the highest average accuracy 
among all PEFT methods. On LLaMA-2-7B, SMoA improves the average score to 82.08, surpassing 
strong baselines such as HiRA (r=32) (81.42) and MeLoRA (r=32, n=2) (79.93). On the more capable 
LLaMA-3-8B, SMoA further advances state-of-the-art methods, reaching an average accuracy of 87.35, 
outperforming HiRA (r=32) (86.73) and DoRA (r=32) (85.20).
These results demonstrate SMoA’s effectiveness in leveraging adapter combinations in different 
spaces to better exploit model capacity and enhance performance within the PEFT framework.

In contrast, for r=16,n=2, SMoA attains an average accuracy of 80\% on LLaMA-2-7B, while using 
substantially fewer trainable parameters than most rank-32 methods, and still outperforming 
or matching several competitive baselines. Similarly, on LLaMA-3-8B, SMoA with r=16 achieves 
an average accuracy of $86.25\%$, demonstrating robust performance despite a reduced intrinsic dimensionality. 
These findings highlight SMoA’s ability to maintain a favorable accuracy–efficiency trade-off, 
confirming its suitability for scalable and resource-conscious model adaptation. 

\subsection{Conversational Task}
Table~\ref{tab:results2} reports a comprehensive comparison of various PEFT methods on 
the CONVAI2 dataset~\cite{Dinan2019TheSC} under two backbone models, LLaMA-2-7B and LLaMA-3-8B. 
Across both backbones, SMoA (r=32, n=2) yields the highest Average score, reaching 47.81 on 
LLaMA-2-7B and 48.42 on LLaMA-3-8B. This represents a clear improvement over strong baselines 
such as LoRA, HiRA, and MeLoRA with comparable parameter budgets. 
Overall, the results demonstrate that SMoA consistently achieves the best or near-best 
performance across most automatic evaluation metrics, validating its effectiveness for dialogue 
generation.

Compared to LoRA and its variants (DoRA, MoRA, SSMLoRA), SMoA consistently delivers higher 
BLEU, METEOR, and ROUGE-L scores. This suggests that SMoA is more effective at capturing both 
surface-level n-gram overlap and longer-range semantic coherence.
Relative to HiRA and MeLoRA, which introduce hierarchical or multi-expert structures, SMoA 
demonstrates superior performance across metrics.
For example, although HiRA achieves strong BERT F1 scores, SMoA surpasses it in BLEU and METEOR.
This suggests that SMoA's multidimensional space adapter approach provides a more balanced 
trade-off between semantic fidelity and generation quality.

\subsection{Mathematical Reasoning Task}
We evaluate the performance of our SMoA method on the GSM8K benchmark using the Llama-3-8B 
model for training. As presented in Table~\ref{tab:gsm8k}, SMoA significantly outperforms 
baseline methods, achieving an accuracy of 72.14\%. This result shows a substantial improvement 
over LoRA (65.89\%), DoRA (66.12\%), and MoRA (67.89\%), demonstrating the effectiveness of 
SMoA in mathematical reasoning tasks. Even with fewer trainable parameters (r=16, n=2), 
SMoA achieves competitive performance, reaching 68.37\%.

These results underscore SMoA's ability to adapt to complex tasks with limited computational 
resources. Its higher rank updates enable the model to maintain a high level of expressiveness 
while optimizing parameter efficiency. Compared to traditional methods like LoRA and its 
variants, SMoA's superior performance indicates the benefits of our approach in enhancing the 
model's capacity for mathematical reasoning.
\begin{figure}[t]
    \centering
    \includegraphics[width=0.85\columnwidth]{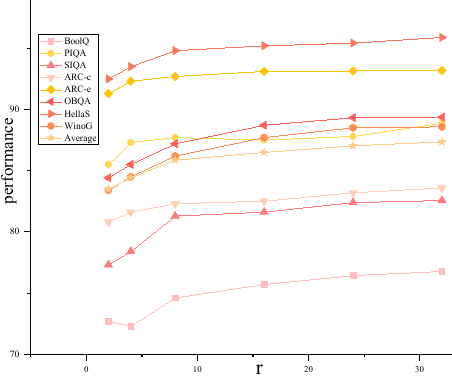}
  \caption{Performance of SMoA across tasks when $r$ increases ($K=2$).}
  \label{fig:performance_rank}
\end{figure}
\begin{table*}[ht]
\begin{adjustbox}{max width=\textwidth, center}
\includegraphics[width=\textwidth]{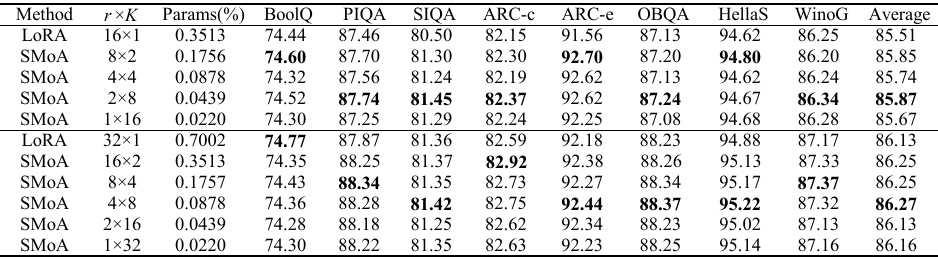}
\end{adjustbox}
\caption{Performance on commonsense reasoning datasets with different equivalent 
ranks ($r\times K$), with the same metrics as in Table~\ref{tab:commonsense}. Bold face indicates best results in terms of the 
corresponding metrics.}
\label{tab:rank_analysis}
\end{table*}
\subsection{Rank Analysis}
We compare the average rank of the update parameter matrix $\Delta W$ for various PEFT methods,
including HiRA, LoRA, MoRA, and SMoA, all of which employ different strategies for 
adapting to new tasks. As shown in Figure~\ref{fig:rank}, SMoA achieves significantly higher ranks for
$\Delta W$ across all values of $r$, indicating its superior ability to perform high-rank adaptation.
Compared to LoRA, MoRA, and MeLoRA, SMoA's rank increases dramatically as $r$ increases, 
suggesting that it is more efficient in capturing task-specific information through 
higher-rank updates. Notably, the rank of $\Delta W$ for HiRA is also considerably higher than that for LoRA and MoRA,
though it does not reach the levels of SMoA.
This result highlights HiRA’s effectiveness in improving the expressiveness of the model 
while maintaining a more parameter-efficient approach. Both HiRA and SMoA demonstrate 
that higher-rank updates correlate with improved performance, as evidenced in Table 1.

In summary, the results emphasize the importance of high-rank adaptation in PEFT methods, 
with SMoA leading in this aspect, achieving substantial improvements over baseline methods, 
and offering a promising direction for enhancing model performance.

\subsection{Impact of $r$}
\label{tab:impactofrank}

Figure~\ref{fig:performance_rank} demonstrates how the performance of SMoA changes with respect 
to the rank parameter $r$ across several commonsense reasoning tasks. As $r$ 
increases from 2 to 32, we observe a consistent and notable improvement in the model’s 
performance, with average accuracy rising from 83.49\% to 87.35\%. This trend highlights the 
impact of increasing the rank $r$ on the model's ability to generalize across diverse reasoning 
tasks.
In particular, tasks such as SIQA, OBQA and WinoG exhibit the most significant performance gains 
as $r$ increases. SIQA, for example, improves by 5.26\% as $r$ rises, emphasizing the 
importance of a higher rank structure for tasks that demand more sophisticated reasoning 
capabilities. Notably, even when $r=8$, SMoA demonstrates a competitive edge, suggesting that 
it efficiently leverages a smaller number of trainable parameters while maintaining strong 
performance, making SMoA a compelling choice even under constrained resources.

\subsection{Analysis of Equivalent Rank}
In this section, we delve into the effect of the subspace configuration on the performance of 
SMoA across different tasks. We conduct experiments with varying configurations of 
$r \times K$. The results are shown in Table~\ref{tab:rank_analysis}.

We have two key observations from the results. First, SMoA consistently achieves superior
or comparable performance across all equivalent rank settings. In Table~\ref{tab:rank_analysis},
SMoA achieves the best performance on most datasets with more than 2 times fewer trainable parameters.
This indicates that the equivalent rank is more important than the number of trainable parameters.
The $2 \times 8$ configuration stands out as the most balanced in terms of both 
performance and parameter efficiency. Achieving an average score of 85.87, it offers competitive 
results without unnecessarily increasing the number of parameters. This highlights the importance 
of selecting a configuration that optimizes both model complexity and computational efficiency.
Ideally, we should search the best equivalent rank settings on different datasets, but setting
$k=2$ is a good choice in most cases.  Second, the optimal equivalent ranks vary across datasets 
and tasks. On ARC-e and HellaS, the optimal performance is generally achieved with $k=2$. In contrast,
for other tasks, a higher value of $k$ such as $4$ or $8$ is a more effective choice.
We think that model sizes and task complexity are the main factors.

\section{Conclusion}
In this paper, we propose a new parameter-efficient fine-tuning method, SMoA, that modulates the features of the original LLM in multiple subspaces, with each subspace learning different features. 
We have theoretically demonstrated that SMoA maintains a higher and flexible rank. 
Results based on 8 PEFT methods and 10 datasets show that SMoA achieves a higher rank
and better performance with fewer trainable parameters on multiple datasets.
\section*{Limitations}

This work has the following limitations. We introduce a hyperparameter $K$, which determines the number of subspaces. The optimal $K$ configuration varies for different datasets, and determining the optimal $K$ configuration requires more experiments and costs.

\bibliography{custom}

@article{touvron2023llama,
  title={Llama: Open and efficient foundation language models},
  author={Touvron, Hugo and Lavril, Thibaut and Izacard, Gautier and Martinet, Xavier and Lachaux, Marie-Anne and Lacroix, Timoth{\'e}e and Rozi{\`e}re, Baptiste and Goyal, Naman and Hambro, Eric and Azhar, Faisal and others},
  journal={arXiv preprint arXiv:2302.13971},
  year={2023}
}

@article{zhang2022opt,
  title={Opt: Open pre-trained transformer language models},
  author={Zhang, Susan and Roller, Stephen and Goyal, Naman and Artetxe, Mikel and Chen, Moya and Chen, Shuohui and Dewan, Christopher and Diab, Mona and Li, Xian and Lin, Xi Victoria and others},
  journal={arXiv preprint arXiv:2205.01068},
  year={2022}
}

@article{achiam2023gpt,
  title={Gpt-4 technical report},
  author={Achiam, Josh and Adler, Steven and Agarwal, Sandhini and Ahmad, Lama and Akkaya, Ilge and Aleman, Florencia Leoni and Almeida, Diogo and Altenschmidt, Janko and Altman, Sam and Anadkat, Shyamal and others},
  journal={arXiv preprint arXiv:2303.08774},
  year={2023}
}

@article{liu2024gpt,
  title={GPT understands, too},
  author={Liu, Xiao and Zheng, Yanan and Du, Zhengxiao and Ding, Ming and Qian, Yujie and Yang, Zhilin and Tang, Jie},
  journal={AI Open},
  volume={5},
  pages={208--215},
  year={2024},
  publisher={Elsevier}
}

@article{hu2022lora,
  title={Lora: Low-rank adaptation of large language models.},
  author={Hu, Edward J and Shen, Yelong and Wallis, Phillip and Allen-Zhu, Zeyuan and Li, Yuanzhi and Wang, Shean and Wang, Lu and Chen, Weizhu and others},
  journal={ICLR},
  volume={1},
  number={2},
  pages={3},
  year={2022}
}

@article{hayou2024lora+,
  title={Lora+: Efficient low rank adaptation of large models},
  author={Hayou, Soufiane and Ghosh, Nikhil and Yu, Bin},
  journal={arXiv preprint arXiv:2402.12354},
  year={2024}
}

@article{jiang2024mora,
  title={Mora: High-rank updating for parameter-efficient fine-tuning},
  author={Jiang, Ting and Huang, Shaohan and Luo, Shengyue and Zhang, Zihan and Huang, Haizhen and Wei, Furu and Deng, Weiwei and Sun, Feng and Zhang, Qi and Wang, Deqing and others},
  journal={arXiv preprint arXiv:2405.12130},
  year={2024}
}

@article{lialin2023relora,
  title={Relora: High-rank training through low-rank updates},
  author={Lialin, Vladislav and Shivagunde, Namrata and Muckatira, Sherin and Rumshisky, Anna},
  journal={arXiv preprint arXiv:2307.05695},
  year={2023}
}

@article{zhuang2024time,
  title={Time-Varying LoRA: Towards effective cross-domain fine-tuning of diffusion models},
  author={Zhuang, Zhan and Zhang, Yulong and Wang, Xuehao and Lu, Jiangang and Wei, Ying and Zhang, Yu},
  journal={Advances in Neural Information Processing Systems},
  volume={37},
  pages={73920--73951},
  year={2024}
}

@inproceedings{huang2025hira,
  title={HiRA: Parameter-efficient hadamard high-rank adaptation for large language models},
  author={Huang, Qiushi and Ko, Tom and Zhuang, Zhan and Tang, Lilian and Zhang, Yu},
  booktitle={The Thirteenth International Conference on Learning Representations},
  year={2025}
}

@inproceedings{liu2021ptuning,
  title={P-Tuning v2: Prompt Tuning Can Be Comparable to Fine-tuning Universally Across Scales and Tasks},
  author={Liu, Xiang and others},
  booktitle={ACL},
  year={2021}
}

@article{xia2024chain,
  title={Chain of lora: Efficient fine-tuning of language models via residual learning},
  author={Xia, Wenhan and Qin, Chengwei and Hazan, Elad},
  journal={arXiv preprint arXiv:2401.04151},
  year={2024}
}

@article{liu2025look,
  title={Look Within or Look Beyond? A Theoretical Comparison Between Parameter-Efficient and Full Fine-Tuning},
  author={Liu, Yongkang and Xu, Xingle and Nie, Ercong and Wang, Zijing and Feng, Shi and Wang, Daling and Li, Qian and Sch{\"u}tze, Hinrich},
  journal={arXiv preprint arXiv:2505.22355},
  year={2025}
}

@inproceedings{lester2021power,
  title={The Power of Scale for Parameter-Efficient Prompt Tuning},
  author={Lester, Brian and Al-Rfou, Rami and Constant, Noah},
  booktitle={Proceedings of the 2021 Conference on Empirical Methods in Natural Language Processing},
  pages={3045--3059},
  year={2021}
}

@article{rebuffi2017learning,
  title={Learning multiple visual domains with residual adapters},
  author={Rebuffi, Sylvestre-Alvise and Bilen, Hakan and Vedaldi, Andrea},
  journal={Advances in neural information processing systems},
  volume={30},
  year={2017}
}

@article{lialin2023stack,
  title={Stack more layers differently: High-rank training through low-rank updates},
  author={Lialin, Vladislav and Muckatira, Sherin and Shivagunde, Namrata and Rumshisky, Anna},
  year={2023}
}

@inproceedings{ren2024melora,
  title={MELoRA: Mini-Ensemble Low-Rank Adapters for Parameter-Efficient Fine-Tuning},
  author={Ren, Pengjie and Shi, Chengshun and Wu, Shiguang and Zhang, Mengqi and Ren, Zhaochun and Rijke, Maarten and Chen, Zhumin and Pei, Jiahuan},
  booktitle={Proceedings of the 62nd Annual Meeting of the Association for Computational Linguistics (Volume 1: Long Papers)},
  pages={3052--3064},
  year={2024}
}

@article{han2024parameter,
  title={Parameter-efficient fine-tuning for large models: A comprehensive survey},
  author={Han, Zeyu and Gao, Chao and Liu, Jinyang and Zhang, Jeff and Zhang, Sai Qian},
  journal={arXiv preprint arXiv:2403.14608},
  year={2024}
}

@article{kopiczko2023vera,
  title={Vera: Vector-based random matrix adaptation},
  author={Kopiczko, Dawid J and Blankevoort, Tijmen and Asano, Yuki M},
  journal={arXiv preprint arXiv:2310.11454},
  year={2023}
}

@inproceedings{liu2024dora,
  title={DoRA: weight-decomposed low-rank adaptation},
  author={Liu, Shih-Yang and Wang, Chien-Yi and Yin, Hongxu and Molchanov, Pavlo and Wang, Yu-Chiang Frank and Cheng, Kwang-Ting and Chen, Min-Hung},
  booktitle={Proceedings of the 41st International Conference on Machine Learning},
  pages={32100--32121},
  year={2024}
}

@inproceedings{wenbatched,
  title={Batched Low-Rank Adaptation of Foundation Models},
  author={Wen, Yeming and Chaudhuri, Swarat},
  booktitle={The Twelfth International Conference on Learning Representations}
}

@inproceedings{yu2025ssmlora,
  title={SSMLoRA: Enhancing Low-Rank Adaptation with State Space Model},
  author={Yu, Jiayang and Zhang, Yihang and Wang, Bin and Lin, Peiqin and Liu, Yongkang and Feng, Shi},
  booktitle={Proceedings of the 2025 Conference of the Nations of the Americas Chapter of the Association for Computational Linguistics: Human Language Technologies (Volume 1: Long Papers)},
  pages={4493--4506},
  year={2025}
}

@inproceedings{zhuang2025come,
  title={Come Together, But Not Right Now: A Progressive Strategy to Boost Low-Rank Adaptation},
  author={Zhuang, Zhan and Wang, Xiequn and Li, Wei and Zhang, Yulong and Huang, Qiushi and Chen, Shuhao and Wang, Xuehao and Wei, Yanbin and Nie, Yuhe and Ma, Kede and others},
  booktitle={42nd International Conference on Machine Learning, ICML 2025},
  year={2025}
}

@inproceedings{hu2023llm,
  title={Llm-adapters: An adapter family for parameter-efficient fine-tuning of large language models},
  author={Hu, Zhiqiang and Wang, Lei and Lan, Yihuai and Xu, Wanyu and Lim, Ee-Peng and Bing, Lidong and Xu, Xing and Poria, Soujanya and Lee, Roy},
  booktitle={Proceedings of the 2023 conference on empirical methods in natural language processing},
  pages={5254--5276},
  year={2023}
}

@inproceedings{papineni2002bleu,
  title={Bleu: a method for automatic evaluation of machine translation},
  author={Papineni, Kishore and Roukos, Salim and Ward, Todd and Zhu, Wei-Jing},
  booktitle={Proceedings of the 40th annual meeting of the Association for Computational Linguistics},
  pages={311--318},
  year={2002}
}

@article{zhang2019bertscore,
  title={Bertscore: Evaluating text generation with bert},
  author={Zhang, Tianyi and Kishore, Varsha and Wu, Felix and Weinberger, Kilian Q and Artzi, Yoav},
  journal={arXiv preprint arXiv:1904.09675},
  year={2019}
}

@article{loshchilov2017decoupled,
  title={Decoupled weight decay regularization},
  author={Loshchilov, Ilya and Hutter, Frank},
  journal={arXiv preprint arXiv:1711.05101},
  year={2017}
}

@article{Cobbe2021TrainingVT,
  title={Training Verifiers to Solve Math Word Problems},
  author={Karl Cobbe and Vineet Kosaraju and Mo Bavarian and Mark Chen and Heewoo Jun and Lukasz Kaiser and Matthias Plappert and Jerry Tworek and Jacob Hilton and Reiichiro Nakano and Christopher Hesse and John Schulman},
  journal={ArXiv},
  year={2021},
  volume={abs/2110.14168},
  url={https://api.semanticscholar.org/CorpusID:239998651}
}

@article{Touvron2023Llama2O,
  title={Llama 2: Open Foundation and Fine-Tuned Chat Models},
  author={Hugo Touvron and Louis Martin and Kevin R. Stone and Peter Albert and Amjad Almahairi and Yasmine Babaei and Niko-lay Bashlykov and Soumya Batra and Prajjwal Bhargava and Shruti Bhosale and Daniel M. Bikel and Lukas Blecher and Cristian Canton Ferrer and Moya Chen and Guillem Cucurull and David Esiobu and Jude Fernandes and Jeremy Fu and Wenyin Fu and Brian Fuller and Cynthia Gao and Vedanuj Goswami and Naman Goyal and Anthony S. Hartshorn and Saghar Hosseini and Rui Hou and Hakan Inan and Marcin Kardas and Viktor Kerkez and Madian Khabsa and Isabel M. Kloumann and Artem Korenev and Punit Singh Koura and Marie-Anne Lachaux and Thibaut Lavril and Jenya Lee and Diana Liskovich and Yinghai Lu and Yuning Mao and Xavier Martinet and Todor Mihaylov and Pushkar Mishra and Igor Molybog and Yixin Nie and Andrew Poulton and Jeremy Reizenstein and Rashi Rungta and Kalyan Saladi and Alan Schelten and Ruan Silva and Eric Michael Smith and R. Subramanian and Xia Tan and Binh Tang and Ross Taylor and Adina Williams and Jian Xiang Kuan and Puxin Xu and Zhengxu Yan and Iliyan Zarov and Yuchen Zhang and Angela Fan and Melissa Hall Melanie Kambadur and Sharan Narang and Aur'elien Rodriguez and Robert Stojnic and Sergey Edunov and Thomas Scialom},
  journal={ArXiv},
  year={2023},
  volume={abs/2307.09288},
  url={https://api.semanticscholar.org/CorpusID:259950998}
}

@inproceedings{Dubey2024TheL3,
  title={The Llama 3 Herd of Models},
  author={Abhimanyu Dubey and Abhinav Jauhri and Abhinav Pandey and Abhishek Kadian and Ahmad Al-Dahle and Aiesha Letman and Akhil Mathur and Alan Schelten and Amy Yang and Angela Fan and Anirudh Goyal and Anthony S. Hartshorn and Aobo Yang and Archi Mitra and Archie Sravankumar and Artem Korenev and Arthur Hinsvark and Arun Rao and Aston Zhang and Aur'elien Rodriguez and Austen Gregerson and Ava Spataru and Baptiste Rozi{\`e}re and Bethany Biron and Binh Tang and Bobbie Chern and Charlotte Caucheteux and Chaya Nayak and Chloe Bi and Chris Marra and Chris McConnell and Christian Keller and Christophe Touret and Chunyang Wu and Corinne Wong and Cristian Canton Ferrer and Cyrus Nikolaidis and Damien Allonsius and Daniel Song and Danielle Pintz and Danny Livshits and David Esiobu and Dhruv Choudhary and Dhruv Mahajan and Diego Garcia-Olano and Diego Perino and Dieuwke Hupkes and Egor Lakomkin and Ehab A. AlBadawy and E I Lobanova and Emily Dinan and Eric Michael Smith and Filip Radenovic and Frank Zhang and Gabriele Synnaeve and Gabrielle Lee and Georgia Lewis Anderson and Graeme Nail and Gr{\'e}goire Mialon and Guanglong Pang and Guillem Cucurell and Hailey Nguyen and Hannah Korevaar and Hu Xu and Hugo Touvron and Iliyan Zarov and Imanol Arrieta Ibarra and Isabel M. Kloumann and Ishan Misra and Ivan Evtimov and Jade Copet and Jaewon Lee and Jan Geffert and Jana Vranes and Jason Park and Jay Mahadeokar and Jeet Shah and Jelmer van der Linde and Jennifer Billock and Jenny Hong and Jenya Lee and Jeremy Fu and Jianfeng Chi and Jianyu Huang and Jiawen Liu and Jie Wang and Jiecao Yu and Joanna Bitton and Joe Spisak and Jongsoo Park and Joseph Rocca and Joshua Johnstun and Joshua Saxe and Ju-Qing Jia and Kalyan Vasuden Alwala and K. Upasani and Kate Plawiak and Keqian Li and Kenneth Heafield and Kevin R. Stone and Khalid El-Arini and Krithika Iyer and Kshitiz Malik and Kuen-ley Chiu and Kunal Bhalla and Lauren Rantala-Yeary and Laurens van der Maaten and Lawrence Chen and Liang Tan and Liz Jenkins and Louis Martin and Lovish Madaan and Lubo Malo and Lukas Blecher and Lukas Landzaat and Luke de Oliveira and Madeline Muzzi and Ma-hesh Pasupuleti and Mannat Singh and Manohar Paluri and Marcin Kardas and Mathew Oldham and Mathieu Rita and Maya Pavlova and Melissa Hall Melanie Kambadur and Mike Lewis and Min Si and Mitesh Kumar Singh and Mona Hassan and Naman Goyal and Narjes Torabi and Niko-lay Bashlykov and Nikolay Bogoychev and Niladri S. Chatterji and Olivier Duchenne and Onur cCelebi and Patrick Alrassy and Pengchuan Zhang and Pengwei Li and Petar Vasi{\'c} and Peter Weng and Prajjwal Bhargava and Pratik Dubal and Praveen Krishnan and Punit Singh Koura and Puxin Xu and Qing He and Qingxiao Dong and Ragavan Srinivasan and Raj Ganapathy and Ramon Calderer and Ricardo Silveira Cabral and Robert Stojnic and Roberta Raileanu and Rohit Girdhar and Rohit Patel and Romain Sauvestre and Ron-nie Polidoro and Roshan Sumbaly and Ross Taylor and Ruan Silva and Rui Hou and Rui Wang and Saghar Hosseini and Sa-hana Chennabasappa and Sanjay Singh and Sean Bell and Seohyun Sonia Kim and Sergey Edunov and Shaoliang Nie and Sharan Narang and Sharath Chandra Raparthy and Sheng Shen and Shengye Wan and Shruti Bhosale and Shun Zhang and Simon Vandenhende and Soumya Batra and Spencer Whitman and Sten Sootla and St{\'e}phane Collot and Suchin Gururangan and Sydney Borodinsky and Tamar Herman and Tara Fowler and Tarek Sheasha and Thomas Georgiou and Thomas Scialom and Tobias Speckbacher and Todor Mihaylov and Tong Xiao and Ujjwal Karn and Vedanuj Goswami and Vibhor Gupta and Vignesh Ramanathan and Viktor Kerkez and Vincent Gonguet and Vir-ginie Do and Vish Vogeti and Vladan Petrovic and Weiwei Chu and Wenhan Xiong and Wenyin Fu and Whit-ney Meers and Xavier Martinet and Xiaodong Wang and Xiaoqing Ellen Tan and Xinfeng Xie and Xuchao Jia and Xuewei Wang and Yaelle Goldschlag and Yashesh Gaur and Yasmine Babaei and Yiqian Wen and Yiwen Song and Yuchen Zhang and Yue Li and Yuning Mao and Zacharie Delpierre Coudert and Zhengxu Yan and Zhengxing Chen and Zoe Papakipos and Aaditya K. Singh and Aaron Grattafiori and Abha Jain and Adam Kelsey and Adam Shajnfeld and Adi Gangidi and Adolfo Victoria and Ahuva Goldstand and Ajay Menon and Ajay Sharma and Alex Boesenberg and Alex Vaughan and Alexei Baevski and Allie Feinstein and Amanda Kallet and Amit Sangani and Anam Yunus and Andrei Lupu and Andres Alvarado and Andrew Caples and Andrew Gu and Andrew Ho and Andrew Poulton and Andrew Ryan and Ankit Ramchandani and Annie Franco and Aparajita Saraf and Arkabandhu Chowdhury and Ashley Gabriel and Ashwin Bharambe and Assaf Eisenman and Azadeh Yazdan and Beau James and Ben Maurer and Benjamin Leonhardi and Po-Yao (Bernie) Huang and Beth Loyd and Beto de Paola and Bhargavi Paranjape and Bing Liu and Bo Wu and Boyu Ni and Braden Hancock and Bram Wasti and Brandon Spence and Brani Stojkovic and Brian Gamido and Britt Montalvo and Carl Parker and Carly Burton and Catalina Mejia and Changhan Wang and Changkyu Kim and Chao Zhou and Chester Hu and Ching-Hsiang Chu and Chris Cai and Chris Tindal and Christoph Feichtenhofer and Damon Civin and Dana Beaty and Daniel Kreymer and Shang-Wen Li and Danny Wyatt and David Adkins and David Xu and Davide Testuggine and Delia David and Devi Parikh and Diana Liskovich and Didem Foss and Dingkang Wang and Duc Le and Dustin Holland and Edward Dowling and Eissa Jamil and Elaine Montgomery and Eleonora Presani and Emily Hahn and Emily Wood and Erik Brinkman and Esteban Arcaute and Evan Dunbar and Evan Smothers and Fei Sun and Felix Kreuk and Feng Tian and Firat Ozgenel and Francesco Caggioni and Francisco (Paco) Guzm{\'a}n and Frank J. Kanayet and Frank Seide and Gabriela Medina Florez and Gabriella Schwarz and Gada Badeer and Georgia Swee and Gil Halpern and Govind Thattai and Grant Herman and Grigory G. Sizov and Guangyi Zhang and Guna Lakshminarayanan and Hamid Shojanazeri and Han Zou and Hannah Wang and Han Zha and Haroun Habeeb and Harrison Rudolph and Helen Suk and Henry Aspegren and Hunter Goldman and Igor Molybog and Igor Tufanov and Irina-Elena Veliche and Itai Gat and Jake Weissman and James Geboski and James Kohli and Japhet Asher and Jean-Baptiste Gaya and Jeff Marcus and Jeff Tang and Jennifer Chan and Jenny Zhen and Jeremy Reizenstein and Jeremy Teboul and Jessica Zhong and Jian Jin and Jingyi Yang and Joe Cummings and Jon Carvill and Jon Shepard and Jonathan McPhie and Jonathan Torres and Josh Ginsburg and Junjie Wang and Kaixing(Kai) Wu and U KamHou and Karan Saxena and Karthik Prasad and Kartikay Khandelwal and Katayoun Zand and Kathy Matosich and Kaushik Veeraraghavan and Kelly Michelena and Keqian Li and Kun Huang and Kunal Chawla and Kushal Lakhotia and Kyle Huang and Lailin Chen and Lakshya Garg and A Lavender and Leandro Silva and Lee Bell and Lei Zhang and Liangpeng Guo and Licheng Yu and Liron Moshkovich and Luca Wehrstedt and Madian Khabsa and Manav Avalani and Manish Bhatt and Maria Tsimpoukelli and Martynas Mankus and Matan Hasson and Matthias Lennie and Matthias Reso and Maxim Groshev and Maxim Naumov and Maya Lathi and Meghan Keneally and Michael L. Seltzer and Michal Valko and Michelle Restrepo and Mihir Patel and Mik Vyatskov and Mikayel Samvelyan and Mike Clark and Mike Macey and Mike Wang and Miquel Jubert Hermoso and Mo Metanat and Mohammad Rastegari and Mun-ish Bansal and Nandhini Santhanam and Natascha Parks and Natasha White and Navy-ata Bawa and Nayan Singhal and Nick Egebo and Nicolas Usunier and Nikolay Pavlovich Laptev and Ning Dong and Ning Zhang and Norman Cheng and Oleg Chernoguz and Olivia Hart and Omkar Salpekar and Ozlem Kalinli and Parkin Kent and Parth Parekh and Paul Saab and Pavan Balaji and Pe-dro Rittner and Philip Bontrager and Pierre Roux and Piotr Doll{\'a}r and Polina Zvyagina and Prashant Ratanchandani and Pritish Yuvraj and Qian Liang and Rachad Alao and Rachel Rodriguez and Rafi Ayub and Raghotham Murthy and Raghu Nayani and Rahul Mitra and Raymond Li and Rebekkah Hogan and Robin Battey and Rocky Wang and Rohan Maheswari and Russ Howes and Ruty Rinott and Sai Jayesh Bondu and Samyak Datta and Sara Chugh and Sara Hunt and Sargun Dhillon and S. Yu. Sidorov and Satadru Pan and Saurabh Verma and Seiji Yamamoto and Sharadh Ramaswamy and Shaun Lindsay and Sheng Feng and Shenghao Lin and Shengxin Cindy Zha and Shiva Shankar and Shuqiang Zhang and Sinong Wang and Sneha Agarwal and Soji Sajuyigbe and Soumith Chintala and Stephanie Max and Stephen Chen and Steve Kehoe and Steve Satterfield and Sudarshan Govindaprasad and Sumit Kumar Gupta and Sung-Bae Cho and Sunny Virk and Suraj Subramanian and Sy Choudhury and Sydney Goldman and Tal Remez and Tamar Glaser and Tamara Best and Thilo Kohler and Thomas Robinson and Tianhe Li and Tianjun Zhang and Tim Matthews and Timothy Chou and Tzook Shaked and Varun Vontimitta and Victoria Ajayi and Victoria Montanez and Vijai Mohan and Vinay Satish Kumar and Vishal Mangla and Vlad Ionescu and Vlad Andrei Poenaru and Vlad T. Mihailescu and Vladimir Ivanov and Wei Li and Wenchen Wang and Wenwen Jiang and Wes Bouaziz and Will Constable and Xia Tang and Xiaofang Wang and Xiaojian Wu and Xiaolan Wang and Xide Xia and Xilun Wu and Xinbo Gao and Yanjun Chen and Ye Hu and Ye Jia and Ye Qi and Yenda Li and Yilin Zhang and Ying Zhang and Yossi Adi and Youngjin Nam and Yu Wang and Yuchen Hao and Yundi Qian and Yuzi He and Zach Rait and Zachary DeVito and Zef Rosnbrick and Zhaoduo Wen and Zhenyu Yang and Zhiwei Zhao},
  year={2024},
  url={https://api.semanticscholar.org/CorpusID:271571434}
}

@article{Dinan2019TheSC,
  title={The Second Conversational Intelligence Challenge (ConvAI2)},
  author={Emily Dinan and Varvara Logacheva and Valentin Malykh and Alexander H. Miller and Kurt Shuster and Jack Urbanek and Douwe Kiela and Arthur Szlam and Iulian Serban and Ryan Lowe and Shrimai Prabhumoye and Alan W. Black and Alexander I. Rudnicky and Jason Williams and Joelle Pineau and Mikhail S. Burtsev and Jason Weston},
  journal={ArXiv},
  year={2019},
  volume={abs/1902.00098},
  url={https://api.semanticscholar.org/CorpusID:59553505}
}

\appendix

\label{sec:appendix}
\section{Datasets}
\label{app:data}
The primary dataset categories evaluated include commonsense reasoning, dialogue generation, 
and mathematical reasoning.
The commonsense reasoning datasets include:
\begin{itemize}[leftmargin=*]
    \item \textbf{BoolQ} is a reading comprehension dataset consisting of naturally occurring 
    yes/no questions paired with Wikipedia passages, designed to evaluate a model’s ability to 
    perform semantic understanding and reasoning over text.
    \item \textbf{PIQA} (Physical Interaction Question Answering) is a commonsense reasoning 
    dataset focused on physical knowledge, where models must choose the more plausible solution 
    to a goal-oriented question involving everyday object interactions and real-world physics.
    \item \textbf{SIQA} (Social Interaction Question Answering) is a commonsense reasoning 
    dataset that evaluates a model’s understanding of social interactions and human intentions, 
    requiring it to select the most appropriate explanation or outcome in everyday social
    situations.
    \item \textbf{ARC-c} (AI2 Reasoning Challenge Set) is a multiple-choice question answering 
    dataset composed of grade-school science questions that require complex reasoning and 
    external knowledge, designed to be difficult for surface-level or retrieval-based methods.
    \item \textbf{ARC-e} (AI2 Reasoning Challenge-Easy Set) is a multiple-choice question 
    answering dataset of grade-school science questions that can typically be answered using
    simple reasoning or direct knowledge retrieval, serving as a baseline subset of the 
    ARC benchmark.
    \item \textbf{OBQA} is a multiple-choice question answering dataset focused on elementary-level 
    science, where answering each question requires combining a small set of provided “open book” 
    science facts with additional commonsense knowledge and reasoning.
    \item \textbf{HellaSwag} is a commonsense reasoning dataset that tests a model’s ability 
    to predict plausible next events in everyday situations by choosing the most realistic 
    continuation of a given context.
    \item \textbf{WinoG} (WinoGrande) is a large-scale commonsense reasoning dataset based on 
    Winograd-style pronoun resolution, designed to evaluate a model’s ability to use 
    contextual and commonsense cues to resolve ambiguous references.
\end{itemize}
The statistical information for the dataset used in the inference attempt is shown in Table ~\ref{tab:data_sta}.

The dialogue generation dataset refers to \textbf{ConvAI2}, which is a dialogue dataset for 
persona-based open-domain conversation, where models must generate or select responses that are 
coherent, engaging, and consistent with a given speaker persona.
Mathematical reasoning dataset refers \textbf{GSM8K}. 
Following the configuration of paper~\citep{huang2025hira}, we use MetaMath as the training corpus 
and GSM8K~\citep{Cobbe2021TrainingVT} as the test dataset.

\section{Training Hyperparameters}
\label{app:hyperparameter}
\begin{table*}[ht]
\begin{adjustbox}{max width=0.85\textwidth, center}
\includegraphics[width=0.85\textwidth]{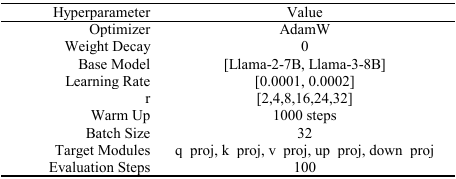}
\end{adjustbox}
\caption{Hyperparameters for SMoA.}
\label{tab:smoa_hyperparameter}
\end{table*}

\begin{table*}[ht]
\begin{adjustbox}{max width=0.85\textwidth, center}
\includegraphics[width=0.85\textwidth]{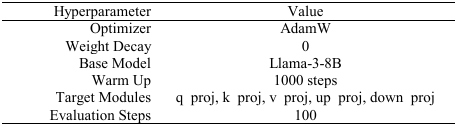}
\end{adjustbox}
\caption{General Hyperparameters Used Across All Experiments.}
\label{tab:gene_hyparameters}
\end{table*}

\begin{table}[ht]
\begin{adjustbox}{max width=\columnwidth, center}
\includegraphics[width=\columnwidth]{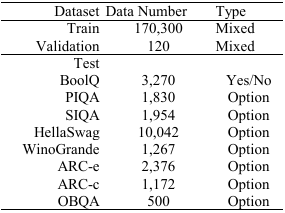}
\end{adjustbox}
\caption{The detailed statistics of commonsense reasoning datasets.}
\label{tab:data_sta}
\end{table}

\end{document}